# Incremental Transductive Learning Approaches to Schistosomiasis Vector Classification

Terence Fusco, Yaxin Bi, Haiying Wang, Fiona Browne
School of Computing and Mathematics
Ulster University
Newtownabbey, United Kingdom
Email: Fusco-T@email.ulster.ac.uk, bi.y@ulster.ac.uk, hy.wang@ulster.ac.uk, f.browne@ulster.ac.uk

*Abstract*—The key issues pertaining to collection of epidemic disease data for our analysis purposes are that it is a labour intensive, time consuming and expensive process resulting in availability of sparse sample data which we use to develop prediction models. To address this sparse data issue, we present novel Incremental Transductive methods to circumvent the data collection process by applying previously acquired data to provide consistent, confidence-based labelling alternatives to field survey research. We investigated various reasoning approaches for semi-supervised machine learning including Bayesian models for labelling data. The results show that using the proposed methods, we can label instances of data with a class of vector density at a high level of confidence. By applying the Liberal and Strict Training Approaches, we provide a labelling and classification alternative to standalone algorithms. The methods in this paper are components in the process of reducing the proliferation of the Schistosomiasis disease and its effects.

## I. INTRODUCTION

A perennial issue that exists concerning application of data mining and machine learning methods for disease classification and prediction is the cost and time involved in the collection of primary field research. In contrast to this we now have access to big data repositories for processing. We are currently capable of accessing vast amounts of unprocessed data mainly due to technology advances for example satellite imagery and feature extraction techniques, but this data requires extensive pre-processing for use with machine learning methods. Processing large quantities of data requires a substantial number of labour hours in addition to specialised knowledge of the subject area which can incur significant financial costs. Viable options that improve efficiency of big data processing by using smaller subsets of data, can be analysed to give a greater understanding of the problem area. This is the problem which we will discuss in this paper. Having the capability to use a snapshot of big data as a whole, for labelling and classification purposes, could expedite the process leading to more efficient results and effective action being taken.

In this paper we investigate density and distribution of the disease vector of the epidemic Schistosomiasis also known as Bilharzia. The solution proposed in this paper is aimed at the problem of correctly labelling and classifying instances of snail density (SD) in order to make accurate predictions of future epidemic disease outbreak. This approach can also be applied to other problem areas suffering from small data sample size. If we can achieve our intended outcome in terms of accurate labelling and classification of snail vector density, then we will reduce the uncertainty of this disease prediction in future with our approach.

Schistosomiasis is a vector-borne disease prevalent in many parts of Africa, Asia and in Brazil in South America. The vector of this disease is the freshwater snail, in our case specifically the Oncomelania Hupensis snail. It is the carrier of the particular strain of the disease known as Schistosomiasis Japonicum [1]. This strain is found mainly in Asia and this reflects our target experiment area which is the Dongting Lake area of Hu'nan province in China [2].

Extensive research has been conducted into the area of Remote Sensing (RS) and Earth Observation (EO) pertaining to epidemic disease spread and possible prevention [3], [4], [5]. The data retrieved from RS and EO methods requires pre-processing for application of machine learning techniques which is time consuming and can be problematic in terms of efficiency. By providing methods which can improve these problems, we hope to make the process much more fluid and remove any time consuming obstacles. It has been shown that utilising labelled data samples can significantly improve classification accuracy while reducing error rates in many different areas such as image recognition [6].

Using the satellite images and environment feature extraction methods, we can make predictions on the class of snail density and distribution which can then be used to gauge the probability of the Schistosomiasis disease spread. Early warning of high risk areas of the disease can alert those living in these municipalities in order for caution to be taken and preventative measures applied. The World Health Organisation as of 2014, states that over 250 million people have been provided with preventative treatment for the disease with over 60 million people having been treated while infected with Schistosomiasis [7].

In this paper we present two Incremental Transductive approaches which address the problem of labelling and classifying data from a limited data pool. The Support Vector Machine classifier (SVM) has been applied to part of this research due to its positive performance in the realm of epidemic disease classification for making disease predictions [8], [9], [10], [11]. The first approach we will look at is the Incremental Transductive Support Vector Machine (ITSVM) which combines an incremental semi-supervised approach with the SVM classifier for labelling and classification purposes, the second is an Incremental Transductive Ensemble Approach (ITEA) which uses a selection of base classifiers that can be specified to a set confidence level which is required either by a single classifier or all classifiers in order for a label to be appointed to the instance . The ITSVM approach is a method of labelling data by means of confidence in the label based on the SVM sequential minimal optimization classification. The incremental element of the approach combined with the Transductive reasoning and SVM confidence threshold setting is conducive to efficient and consistent data labelling and classification for our unlabelled data. The confidence threshold setting in any case is dependable on the requirement of the test area being analysed. If it is decided that a benchmark confidence level is required for the labelling of a particular set of data, then that is what shall be selected. For the purposes of scientific fairness during experiments in this paper, the data from each year was normalised and split equally in 3 and 5 category groupings of SD classes. On completion of this research, we may decide on which category distribution we will use going forward for classification and prediction purposes.

The consecutive sections of the paper are laid out as follows: Firstly we have Section II on Related work which covers some labelling techniques and their application and results in relevant areas. Then in Section III we detail the environment features used in the experiments and discuss some of the methods used accompanied by equations and diagrams to describe the labelling process. In Section IV we provide the results of methods used in this research with corresponding tables for each year's results. We then have some discussion in Section V followed by a summary in the Section VI conclusion.

## II. Related Work

Research concerned with healthcare and epidemic disease control has benefited from satellite technology and artificial intelligence applications increasingly in recent years [12], [13], [14]. Much work has been carried out with the objective of predicting or mapping the risk of future disease outbreak; which can prove to be of critical importance especially for those in remote areas [15], [16]. Having access to relevant data for experiment purposes is integral to conducting high quality research which we can then use to compare methods for disease detection and prediction [17].

The fields of data mining and machine learning application using sparse data has yielded many varying solutions including methods for improving classification performance and accurately labelling classes [18]. Semi-supervised learning methods for labelling of data are becoming increasingly complex and can provide promising results when utilising a sparse sample dataset [19]. Transductive Reasoning is a form of semi-supervised learning which in our case enables us to label the SD values from a dataset using specifically selected and observed training data. Studies have shown that Transductive reasoning for classification purposes can achieve higher accuracy than Inductive reasoning models when applied together with SVM in the realms of Remote Sensing and Bioinformatics [20], [21]. These studies form part of the rationale for this research as we are building on previous theory and applying the process to our epidemic prediction models.

In this paper we use different data combinations for training and testing purposes with specific cases for example, when testing the confidence levels for labelling of data from different years. Transductive learning models help us understand the problem area by providing information on confidence levels achievable when labelling classes of data. This helps us recognize those instances which classifiers have most difficulty in providing a label. The implementation of the SVM classifier gives us a confidence rating in our ITSVM method when labelling the SD data and we use this for consistency and have confidence in the label in accordance with whichever threshold we select. The incremental methods used during this research include using classification algorithms such as SVM, Naive Bayes and J48 as they provide well established approaches and variation when we are seeking to get a consensus on our labelling confidence. Our SD data has been normalised and each class given equal weighting from 0-1 as is standard in transductive research of this nature [22].

## III. Methods and Materials

The following section details the dataset information used in experiments for this work as well as describing how the proposed applications are applied to the data. By experimenting with the confidence and passive threshold levels, we can assess the optimum level at which each dataset can perform in terms of the labelling and SD classification of the data.

The pseudocode for the LTA and STA algorithms used in the ITEA can be seen in Algorithm 1: Where $L$ is the set of labelled instances, $U$ is the set of unlabelled instances, $l$ is a labelled instance, and $u$ is an unlabelled instance.

We then have n as the number of iterations with no labelled instance, $\gamma$ is the confidence threshold for classifiers and $\phi$ is the number of passive iterations that are permitted without any labelling taking place before ending the process.

### A. Experimental Data

The dataset we have access to is limited in terms of the number of instances (180) due to the time and resources required to carry out the field survey data from the Dongting Lake in China over the spatial and temporal parameters. The common environmental attributes to each year of field research were used for continuity and comparison of classification results. The common attributes are as follows: Tasselled Cap Brightness (TCB), Tasselled Cap Greenness (TCG), Tasselled Cap Wetness (TCW), Modified Normalised Difference Water Index (MNDWI), Normalised Difference Moisture Index (NDMI), Normalised Difference Vegetation Index (NDVI) and Normalised Difference Water Index (NDWI). By using this limited dataset and those attributes common to all sets, we can identify the extent to which we can attain highly accurate

**Algorithm 1** ITEA

```
 1: procedure LTA(L, U, l, u, n, γ, φ)
 2:     while U ≠ ∅ ∨ n ≥ φ do
 3:         n = 0
 4:         TrainClassifiers
 5:         for u ∈ U do
 6:             if P₁ ≥ γ ∨ P₂ ≥ γ ∨ P₃ ≥ γ then
 7:                 L ← u
 8:                 n = 0
 9:             else
10:                 n + 1
11:             end if
12:         end for
13:     end while
14: end procedure

15: procedure STA(L, U, l, u, n, γ, φ)
16:     while U ≠ ∅ ∨ n ≥ φ do
17:         n = 0
18:         TrainClassifiers
19:         for u ∈ U do
20:             if P₁ ≥ γ ∧ P₂ ≥ γ ∧ P₃ ≥ γ then
21:                 L ← u
22:                 n = 0
23:             else
24:                 n + 1
25:             end if
26:         end for
27:     end while
28: end procedure
```

classification when changing the confidence threshold and the passive threshold.

The rationale behind our incremental methods is to select existing well performing algorithms such as Naive Bayes, SVM, J48, Multi-Layer Perceptron etc. and modify the labelling and classification processes by including an incremental element to the models. By applying an incremental approach we aim to improve the accuracy of labels provided and resulting classification compared to the traditional SVM alone. The formalisation for the proposed method can be seen in Equation 1.

In this equation we assume that an instance $i$ is a vector of real number attributes $a$ resulting in $i := (a_1, a_2, ..., a_n)$. Suppose we have a collection of classifiers $K$ and a labelling threshold $\gamma$ which the probability of the class must exceed to be labelled. Then for each instance $i$ in the unlabelled data and for classifier $k \in K$ we perform Equation 1, where $p(c_m|i)$ is the probability of the class $c_m$ from the set of classes $C$ given an instance $i$ generated by a classifier $k$.

$$p(c|i) = \underset{c_m \in C}{argmax} \begin{cases} p(c_m), & if \quad \exists c_m \in C: \quad p(c_m) \geq \gamma \\ 0, & otherwise \end{cases} \quad (1)$$

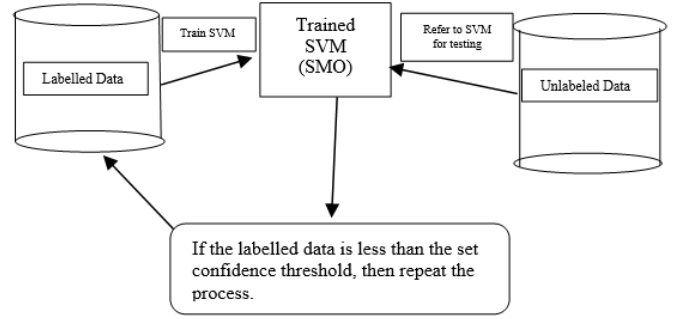

Fig. 1: ITSVM Process

In Equation 2 the Label is assigned to the class if and only if the probability of the class of the instance is greater than or equal to the probability of the class given the instance.

$$L = c_i \quad iff \quad p(c_i) \geq p(c|i) \quad (2)$$

The approaches used in this paper were applied to label a selection of data that was collected using field survey research and remotely sensed image extraction of environment features. Experiments have been carried out on each of our year's data which ranged from 2003 to 2009, and was provided by our partners at the European Space Agency and the Academy of Opto-Electronics in China. All classifiers used in these experiments were applied with the default settings in the interests of equality for comparative analysis.

*B. Incremental Transductive Support Vector Machine*

The SVM classifier has shown to perform well in comparison with traditional algorithms in the area of epidemic disease vector classification [10]. By using this classifier we can hope to attain consistent and highly accurate classification of SD. The Sequential Minimal Optimization implementation of SVM uses an iterative approach by dividing the data problem at hand into subsets of data which enables more efficient problem solving. This approach provides us with levels of confidence when we aim to label our data using the ITSVM method. By implementing this incremental method, we can enhance the training potential of our data by adding newly labelled SD instances to the training process which then repeats until all instances are labelled to the selected confidence threshold level.

The first of our novel applications is the Incremental Transductive Support Vector Machine (ITSVM) method which is shown in Fig. 1. It uses the SVM algorithm by setting a confidence level at which the unlabelled data can be assigned a class label and then when any instances meet that level of confidence, they are then added to the now new dataset to start the process again until the set passive threshold is met or the process no longer is capable of assigning labels to the set level.

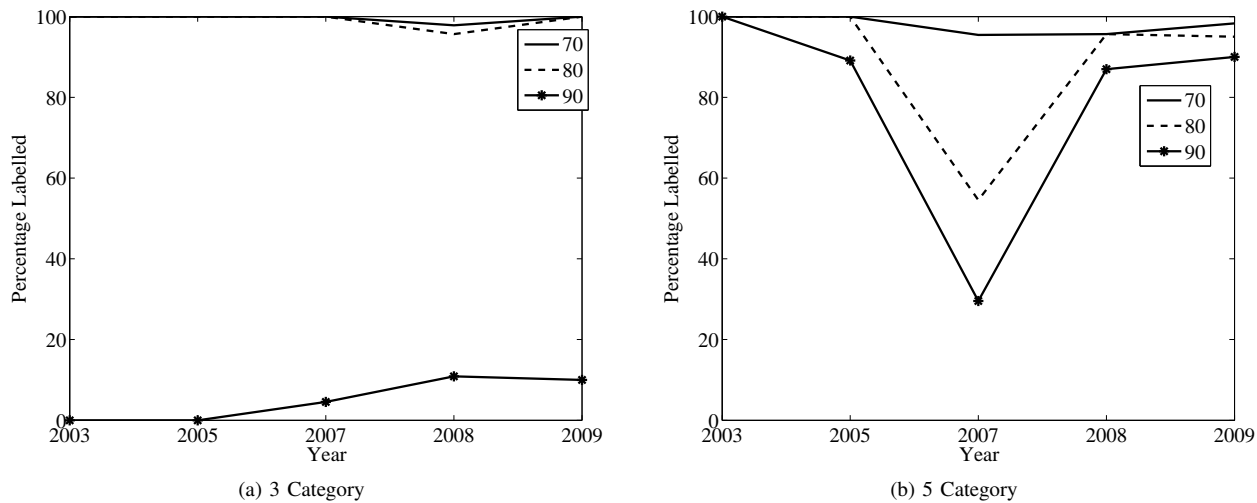

(a) 3 Category      (b) 5 Category

Fig. 2: Liberal Training Approach

*C. Incremental Transductive Ensemble Approach*

The second novel approach we have applied is an Incremental Transductive Ensemble Approach (ITEA). This approach uses a selection of classifiers to improve the labelling process and classification of the data instances. Using multiple classifiers has been shown to improve on performance in related biomedical research fields [23] so we have applied this knowledge to experiments with our ITEA methods. Contained in the ITEA we have the *liberal training approach* (LTA) and *Strict Training Approach* (STA). The ITEA involves a similar process to ITSVM but differing in the multi-classifier selection of the process which applies multiple classifiers to train and label our data for classification. By implementing this approach we have selected the labels of data to be applied in two ways: - The *LTA* requires only one of the selected algorithms to meet the set threshold level before labelling the class. With this approach an ensemble of algorithms is selected, which in our case we have used SVM, Naive Bayes and J48, once any one of the algorithms can apply a label with the confidence threshold level, then the label is assigned. If more than one of the algorithms meet the level but with different labels, then the one which has the highest confidence in the label will be selected. This approach can be beneficial if the classification and data labelling application requires the set confidence of a label to be met by any of the select algorithms for its required purpose. We can see in Fig. 2 that with 3 categories of SD class, we can achieve a higher percentage of labelled data instances when compared with 5 categories. Applying our *LTA* with 5 categories performs at lower levels particularly in year 2007 which is something we address in this paper and plan to give more consideration to when going forward with this research.

The *STA* requires that each of the selected algorithms must concur on the set confidence level in order to provide a label to the class. Unless the collective algorithms reach the confidence level, then there will be no label assigned. This method is particularly of interest when classification is critical to an applied field. The STA is best applied to data which requires a high level of confidence for any labels to be provided to the set considering that the newly labelled data will be included in the training set for classification purposes and this may have serious implications for certain datasets depending on the domain that they are applied to. The high levels of confidence used in these methods are obtained by using small development sets in order to set the threshold. Using this method, it is possible that the data will remain unlabelled if the confidence threshold is set to a higher level than can be achieved by all relevant classifiers. This consequence of unlabelled data instances should only improve the probability of higher classification accuracy with each individual dataset. In Fig. 3, we can see that while both category labelled percentage levels correlate to an extent, there still exists a reduction in labelling with 5 categories in comparison to 3 SD categories.

## IV. RESULTS

Using our Incremental Transductive methods, we can adjust the confidence and passive thresholds to suit our specification on a particular dataset. The set confidence threshold details how positive the method (in this case SVM) is in the label that it has provided while the passive threshold is a measure to limit how many attempts will be made to label the data before the approach declares the instance to remain unlabelled. The confidence threshold ranges from 0-1 with 1 being 100 percent confidence. For the ITSVM experiment, the confidence was initially set to 0.8 and the passive threshold was set at 10 iterations. The application of this ITSVM method can provide us with a model that can give a consistent standard of semi-supervised labelling at any given selected confidence threshold while additionally pinpointing the instances which are found to be most difficult to label for further inspection.

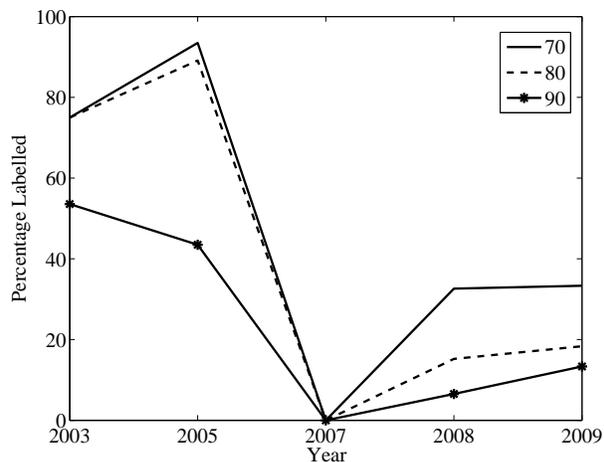
(a) 3 Category

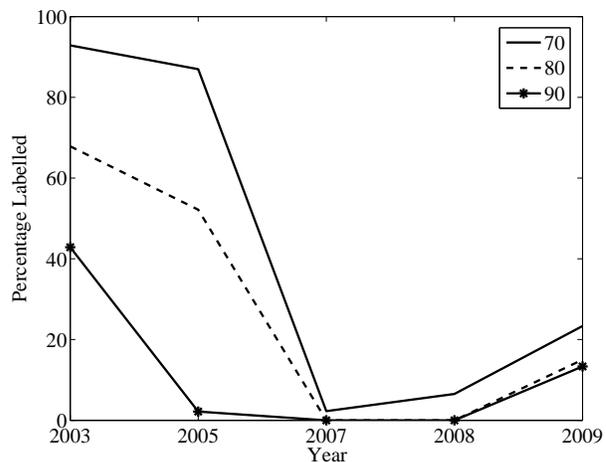
(b) 5 Category

Fig. 3: Strict Training Approach

TABLE I: ITSVM Comparison

| | | | 3CAT | | 5CAT | |
|---|---|---|---|---|---|---|
| Train | Test | Num.Inst. | Inst.Lbled | Iters. | Inst.Lbled | Iters. |
| 2003 | 2005 | 46 | 42 | 7 | 46 | 6 |
| 2003 | 2007 | 44 | 44 | 6 | 31 | 4 |
| 2003 | 2008 | 46 | 40 | 8 | 42 | 4 |
| 2003 | 2009 | 60 | 49 | 5 | 42 | 5 |
| 2005 | 2003 | 28 | 26 | 5 | 12 | 3 |
| 2005 | 2007 | 44 | 40 | 5 | 31 | 4 |
| 2005 | 2008 | 46 | 38 | 5 | 45 | 6 |
| 2005 | 2009 | 60 | 60 | 7 | 59 | 8 |
| 2007 | 2003 | 28 | 26 | 5 | 0 | 1 |
| 2007 | 2005 | 46 | 45 | 7 | 0 | 1 |
| 2007 | 2008 | 46 | 45 | 6 | 0 | 1 |
| 2007 | 2009 | 60 | 60 | 6 | 0 | 1 |
| 2008 | 2003 | 28 | 13 | 3 | 0 | 1 |
| 2008 | 2005 | 46 | 2 | 1 | 0 | 1 |
| 2008 | 2007 | 44 | 34 | 6 | 0 | 1 |
| 2008 | 2009 | 60 | 27 | 3 | 0 | 1 |
| 2009 | 2003 | 28 | 19 | 4 | 18 | 4 |
| 2009 | 2005 | 46 | 45 | 5 | 45 | 5 |
| 2009 | 2007 | 44 | 44 | 6 | 29 | 6 |
| 2009 | 2008 | 46 | 44 | 6 | 37 | 7 |

TABLE II: ITSVM Benchmarking Table

| 70% Conf. | 2003 | 2005 | 2007 | 2008 | 2009 |
|---|---|---|---|---|---|
| *ITSVM 03* | **92.31** | **84.78** | **59.09** | 29.55 | 47.17 |
| *WEKA 03* | 85.19 | **84.78** | 54.55 | 56.52 | 66.67 |
| *ITSVM 05* | 38.46 | **86.96** | 11.63 | 38.1 | 8.33 |
| *WEKA 05* | 85.19 | 84.78 | 54.55 | 56.52 | 66.67 |
| *ITSVM 07* | 84.62 | 11.11 | **64.71** | 46.67 | **66.67** |
| *WEKA 07* | 85.19 | 84.72 | 54.55 | 56.52 | **66.67** |
| *ITSVM 08* | 71.43 | 78.26 | **78.38** | **77.27** | 61.9 |
| *WEKA 08* | 85.19 | 84.78 | 54.55 | 56.52 | 66.67 |
| *ITSVM 09* | 76.19 | 84.44 | **56.82** | **57.78** | **86.11** |
| *WEKA 09* | 85.19 | 84.78 | 54.55 | 56.52 | 66.67 |

*A. ITSVM Results*

The results in Table I were recorded from data which was classified for 3 and 5-point categories of low, medium and high compared with v.low, low, medium, high and v.high SD.

Table I shows the first application of the ITSVM with 3-point and 5-point category classification on our data with the number of instances labelled and iterations required for each year training and testing combination. We can see the labelling success that is possible when using a standard confidence threshold of 0.8 for labelling the SD class data. The key on each line graph shows the line characteristics for 70, 80 and 90 percent set confidence levels and the data lines depict the results for each year with corresponding confidence threshold and percentage labelled. In terms of benchmarking results when compared with the standard SVM using the Weka workbench software, we conducted tests with each year of data and differing confidence thresholds to assess the performance of our methods. The results of the ITSVM outperformed the standard Weka SVM application in terms of accuracy in many cases with some instances having equal classification accuracy. A snapshot of our results using a 70 percent confidence threshold are shown in Table II.

*B. ITEA Results*

It is evident in some cases that using an alternate year of training data to label the classes of an existing year of data can have positive results as shown in Table II the case of training data from 2003 to label the testing data from 2007. When we compare this labelling of the 2003/07 combination to training and testing from 2007, we can see that 2003/07 is more favourable as it is capable of assigning the entire 44 instances with a label which on this inconsistent dataset can be perceived as a positive result. Indeed, when assessing the entire Table, we can see that using year 2003 as a training set is on average the most consistent performing training option for labelling the data using this selected confidence level and 3-point classification framework.

TABLE III: 3CAT ITEA Comparison

|  |  |  | 3CAT LTA | | 3CAT STA | |
|---|---|---|---|---|---|---|
| Training | Confidence | Total Inst. | Lbl % | Remaining Inst. | Lbl % | Remaining Inst. |
| 2003 | 70 | 28 | 100 | 0 | 75 | 7 |
| 2005 | 70 | 46 | 100 | 0 | 93.48 | 3 |
| 2007 | 70 | 44 | 100 | 0 | 0 | 44 |
| 2008 | 70 | 46 | 97.83 | 1 | 32.61 | 31 |
| 2009 | 70 | 60 | 100 | 0 | 33.33 | 40 |
| 2003 | 80 | 28 | 100 | 0 | 75 | 7 |
| 2005 | 80 | 46 | 100 | 0 | 89.13 | 5 |
| 2007 | 80 | 44 | 100 | 0 | 0 | 44 |
| 2008 | 80 | 46 | 95.65 | 2 | 15.22 | 39 |
| 2009 | 80 | 60 | 100 | 0 | 18.33 | 49 |
| 2003 | 90 | 28 | 0 | 28 | 53.57 | 13 |
| 2005 | 90 | 46 | 0 | 46 | 43.48 | 26 |
| 2007 | 90 | 44 | 4.55 | 42 | 0 | 44 |
| 2008 | 90 | 46 | 10.87 | 41 | 6.52 | 43 |
| 2009 | 90 | 60 | 10 | 54 | 13.33 | 52 |

TABLE IV: 5CAT ITEA Comparison

|  |  |  | 5CAT LTA | | 5CAT STA | |
|---|---|---|---|---|---|---|
| Training | Confidence | Total Inst. | Lbl % | Remaining Inst. | Lbl % | Remaining Inst. |
| 2003 | 70 | 28 | 100 | 0 | 92.86 | 2 |
| 2005 | 70 | 46 | 100 | 0 | 86.96 | 6 |
| 2007 | 70 | 44 | 95.45 | 2 | 2.27 | 43 |
| 2008 | 70 | 46 | 95.65 | 2 | 6.52 | 43 |
| 2009 | 70 | 60 | 98.33 | 1 | 23.33 | 46 |
| 2003 | 80 | 28 | 100 | 0 | 67.86 | 9 |
| 2005 | 80 | 46 | 100 | 0 | 52.17 | 22 |
| 2007 | 80 | 44 | 54.55 | 20 | 0 | 44 |
| 2008 | 80 | 46 | 95.65 | 2 | 0 | 46 |
| 2009 | 80 | 60 | 95 | 3 | 15 | 51 |
| 2003 | 90 | 28 | 100 | 0 | 42.86 | 16 |
| 2005 | 90 | 46 | 89.13 | 5 | 2.17 | 45 |
| 2007 | 90 | 44 | 29.55 | 31 | 0 | 44 |
| 2008 | 90 | 46 | 86.96 | 6 | 0 | 46 |
| 2009 | 90 | 60 | 90 | 6 | 13.33 | 52 |

## C. Cumulative Training Approach ITSVM Results

In contrast to the 3-point classification scale in Table I, we can identify that labelling success is much less frequent with the 5-point scale. As can be seen with Table V, there is only one data combination that is capable of labelling all instances at this confidence threshold which is the 2003/05 combination. This is compared with the 3-point classification in Table I which provides five entire classification labelling of all instances. These results were to be expected considering the additional classes that are possible for labelling. In addition to running experiments on both the 3 and 5 point scale for labelling purposes, we have analysed the effect of the ITSVM labelling method on our prior method of a Cumulative Training Approach or CTA which combines multi-years training data for testing against individual years. In this case for labelling the instances, we have applied our CTA to both components of 3 and 5 point category classification. It is the case that only one of each of the 3 and 5 point methods have been able to label all instances in the set. By using this approach to labelling with the ITSVM, we have achieved a lower performance yet more balance between the labelling of both category brackets. There is also an instance which neither of the categories could label more than one instance namely the training data of 2003/05/07/08 for testing 2009 which is worth considering for further experiment as to why the data is unable achieve a higher labelling success rate in this set. The results in Table V were recorded from data which are classified in 5 categories of very low, low, medium, high and very high SD.

TABLE V: CTA ITSVM

| Train | Test | Num.Inst. | Inst.Lbled | Inst.Unlbled | Iters. |
|---|---|---|---|---|---|
| | | 3 CATEGORY SD | | | |
| 2003/05 | 2007 | 44 | 19 | 25 | 4 |
| 07/05/2003 | 2008 | 46 | 46 | 0 | 8 |
| 2003/05/07/08 | 2009 | 60 | 1 | 59 | 1 |
| 08/07/2005 | 2009 | 60 | 9 | 51 | 2 |
| 09/08/2007 | 2003 | 28 | 11 | 17 | 2 |
| 2008/09 | 2005 | 46 | 43 | 3 | 6 |
| | | 5 CATEGORY SD | | | |
| 2003/05 | 2007 | 44 | 7 | 37 | 2 |
| 07/05/2003 | 2008 | 46 | 12 | 34 | 3 |
| 2003/05/07/08 | 2009 | 60 | 1 | 59 | 1 |
| 08/07/2005 | 2009 | 60 | 0 | 60 | 1 |
| 09/08/2007 | 2003 | 28 | 1 | 27 | 1 |
| 2008/09 | 2005 | 46 | 46 | 0 | 6 |

## V. DISCUSSION

We can see from analysis of each of the graphs and tables provided in this paper that during the years 2007 and 08, the labelling and classification methods performed poorly when compared with the alternate remaining years of data for example 2003 and 2005. This is, we believe due to adverse weather conditions and climate issues present over these recorded years which has a strong impact on the epidemic disease study area for SD and distribution [24], [25] . It is indicative of these weather conditions that we had partially complete data in year 2007, and as a result of this, negative correlation in year 2008. The effect of adverse or extreme weather conditions having an impact on our prediction methods is a subject for much consideration [26]. We will conduct further research into how we can approach the changes in environment feature values that derive from these conditions as they will be an issue going forward. Satellite imagery can be affected when applying image extraction during bad weather conditions and therefore this problem is an ongoing one that we need to address.

## VI. CONCLUSION

We have presented in this paper our novel ITSVM approach which proposes an incremental semi-supervised labelling and classification method to our problem of epidemic disease vector prediction. The next stage of development to the ITSVM which is the novel ITEA, provides the LTA and STA paradigms in which we can set parameters for labelling and classification with a variety of base classifiers to a selected confidence threshold level. When comparing ITSVM with the standard SVM approach, we have seen a quarter of all instances in our collective dataset have been improved in terms of classification accuracy. This together with the many equalled accuracy levels render the ITSVM a viable option to implement or incorporate when classifying data of this nature for Schistosomiasis classification and prediction purposes. The ITEA methods have shown to be competitive when labelling SD data with the LTA performing well in many cases whereas with STA we see varied results which require further assessment and analysis. It is evident from the ITSVM approach that we can use our incremental method for accurate labelling and improve classification of the SD class when applied to this epidemic disease problem area. The SVM approach is initially capable of providing classification with confidence levels based on the user selection while the incremental element helps to increase the training potential of the data by including the newly labelled instances to the data pool. In summary, the findings produced from applying the approaches presented in this paper show that using the ITSVM method we can improve on the standard inductive Weka SVM implementation on around 25 percent of our data with many more instances yielding equal classification accuracy. The ITEA Liberal and Strict training approaches provide us with competitive labelling as well as poor and inconsistent performance which we will consider going forward with this research by investigating the particular instances which prove most difficult to label. The next stage of this research will involve analysing these unlabelled data instances and trying to improve the precision of the labelling and classification process. These methods are components of the overarching data amplification techniques that compile our Cumulative Training Paradigm which we are aiming to apply as a framework when faced with limited data that we want to utilise to the maximum potential for training and testing purposes in this research area.

## VII. ACKNOWLEDGEMENTS

This work is partially supported by the Dragon 3 programme, a co-operation between the European Space Agency and the Ministry of Science and Technology of China. The authors would also like to acknowledge the Chinese partners at the Academy of Opto-Electronics, Chinese Academy of Sciences for making this data available for our research.